\documentclass[runningheads]{llncs}

\usepackage{eccv}

\usepackage{eccvabbrv}

\usepackage{graphicx}
\usepackage{booktabs}
\usepackage{wrapfig}
\usepackage{amssymb}
\usepackage{amsmath}
\usepackage{booktabs}
\usepackage{colortbl}
\usepackage{arydshln,hhline}
\usepackage{bm}

\usepackage[accsupp]{axessibility}  

\usepackage{hyperref}

\usepackage{orcidlink}

\begin{document}

\title{SiGMA: Sign-Guided Merging and Adaptation for Multimodal Continual Instruction Tuning} 

\titlerunning{SiGMA}


\author{Keonhee Park\orcidlink{0009-0008-3654-812X} \and
Gunhee Kim\orcidlink{0000-0002-9543-7453}} 

\authorrunning{Keonhee park et al.}

\institute{Seoul National University \\
\email{keonhee.park@vision.snu.ac.kr, gunhee@snu.ac.kr} 
}

\maketitle

\begin{abstract}
  Multimodal Continual Instruction Tuning (MCIT) is crucial for adapting Multimodal Large Language Models (MLLMs) to evolving a sequence of downstream tasks. 
  Prior methods mostly utilize Mixture-of-Experts or expansion–merge approach, primarily focusing on catastrophic forgetting, yet they still suffer from negative interference during inference, where newly learned updates overwrite useful prior knowledge and degrade overall performance. 
  To address this, we propose \textbf{SiGMA} (\textbf{Si}gn-\textbf{G}uided \textbf{M}erging and \textbf{A}daptation), a simple yet effective framework that mitigates negative interference with two components: sign-guided adaptive tuning during training and sign-guided merging at inference. Sign-guided adaptive tuning reduces collisions with past knowledge and learns the current task with minimal drift, mitigating severe forgetting. Sign-guided merging further improves consolidation by selectively scaling salient parameters to preserve and amplify useful task-specific knowledge. Experiments on UCIT and DCL benchmarks show that SiGMA significantly reduces negative interference and outperforms state-of-the-art MCIT methods. Our code is available at \href{https://github.com/pgh2874/SiGMA-Multimodal-Continaul-Instruction-Tuning}{SiGMA}.

  
\end{abstract}

\section{Introduction}
\label{sec:intro}
Recently, Multimodal Large Language Models (MLLMs)~\cite{liu2023visual,chen2024internvl,achiam2023gpt,liu2024improved} have gained significant attention for their strong instruction-following ability and effective vision–language alignment, achieving impressive performance across diverse tasks \cite{lumathvista,tang2023vistext,li2024monkey}. MLLMs are typically trained in pre-training followed by instruction tuning; pre-training learns vision–text aligned representations, while instruction tuning enables the model to follow user instructions for response generation. This training allows for zero-shot capability on unseen instructions. However, such zero-shot ability often fails to meet domain-specific needs, making additional instruction tuning necessary to adapt MLLMs to downstream tasks.

Multimodal Continual Instruction Tuning (MCIT) \cite{guo-etal-2025-hide,zeng2025modalprompt,wang2023orthogonal,chen2024coin,huai2025cl}  enables MLLMs to continually acquire novel knowledge while preserving previously learned knowledge. One key well-known challenge in MCIT is catastrophic forgetting~\cite{mccloskey1989catastrophic}, where learning new tasks significantly degrades performance on previously learned tasks. In the context of MLLMs, catastrophic forgetting can manifest as the loss of previously learned representation knowledge and degradation of instruction-following capability. As a result, the model may generate responses with incorrect content and fail to comply with user-specified constraints.

\begin{figure}[t]
  \centering
  \includegraphics[width=\linewidth]{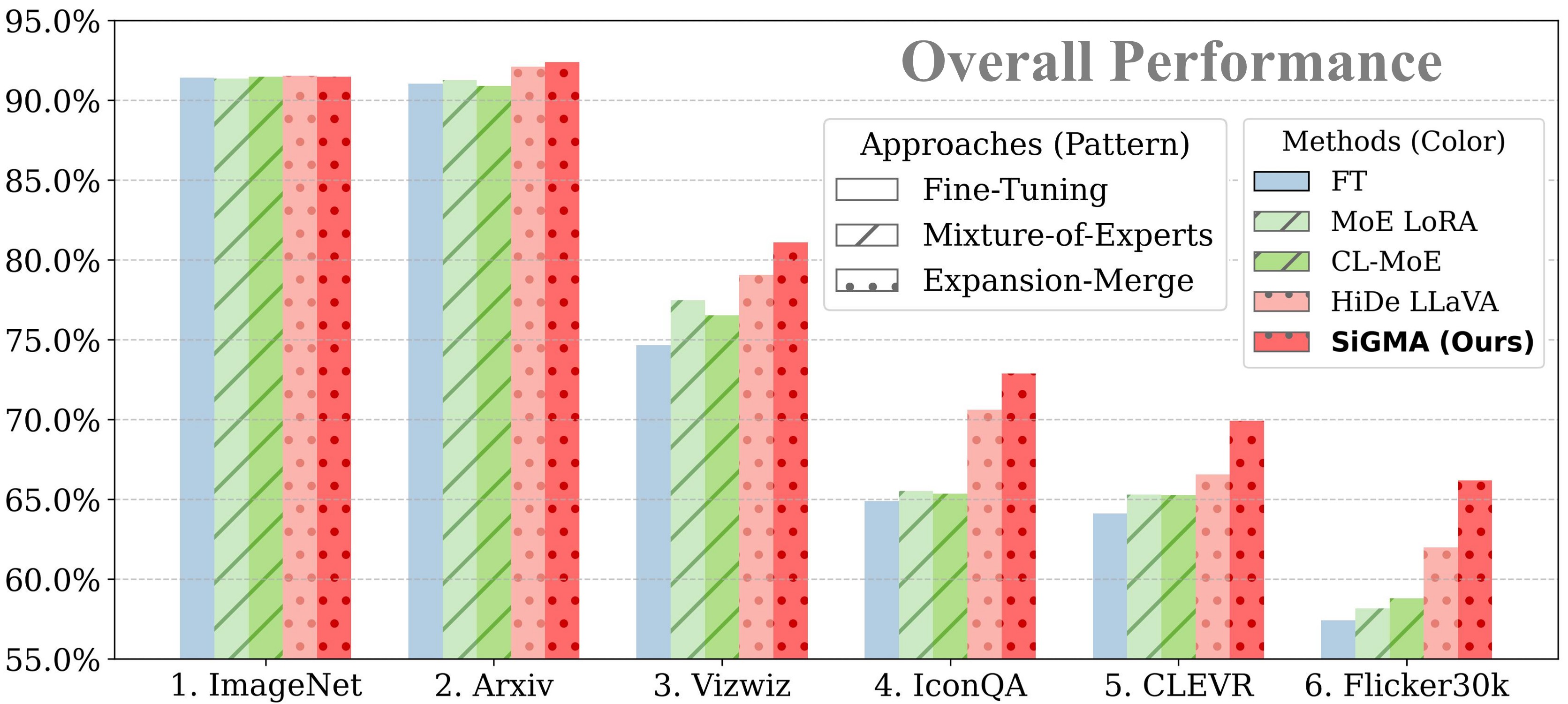}
  \caption{
  Comparison of overall performance on UCIT benchmark. We report a holistic metric over the continual instruction tuning process, computed as the average accuracy on all previously seen tasks after training each task. While existing MoE and expansion–merge approaches gradually suffer from performance degradation due to negative interference and severe forgetting, SiGMA, which explicitly addresses negative interference, consistently maintains high overall performance.
  }
  \label{fig:motiv}
\end{figure}

Conventional MCIT methods typically rely on LoRA~\cite{hu2022lora} to adapt MLLMs to new tasks while preserving previously learned knowledge. Most prior work mitigates catastrophic forgetting through either Mixture-of-Experts (MoE)~\cite{chen2024coin,huai2025cl} or expansion–merge strategies~\cite{guo-etal-2025-hide,wang2023orthogonal}. MoE distributes learned knowledge across experts to reduce forgetting, whereas expansion–merge strategies train a task-specific LoRA for each task, freeze it after training, and merge the learned LoRAs at inference to consolidate task-specific knowledge. 
As shown in~\Cref{fig:motiv}, despite their effectiveness in mitigating forgetting, both approaches remain vulnerable to negative interference~\cite{aljundi2019online}, where newly learned knowledge may overwrite earlier knowledge and degrade comprehensive performance throughout the continual learning process. In consequence, beyond preventing catastrophic forgetting, mitigating negative interference is essential for robust MCIT.

To this end, we propose the SiGMA (\textbf{Si}gn-\textbf{G}uided \textbf{M}erging and \textbf{A}daptation) framework that combines two simple yet effective strategies: \textit{sign-guided adaptive tuning} during training and \textit{sign-guided merging} at inference. The core idea of SiGMA is to decouple LoRA weights into general (task-invariant) and specific (task-variant) subspaces based on the parameter-wise sign pattern. From our empirical observation,  parameters that share the same sign across tasks can be treated as a general subspace, whereas those with differing signs can be treated as a specific subspace. 
During training, sign-guided adaptive tuning reduces collisions with previously learned knowledge and adapts the current LoRA weights by leveraging both the pre-trained MLLM and the historical general subspace, encouraging task learning with minimal drift. In inference, sign-guided merging integrates multiple LoRA weights by selecting salient parameters from the specific subspace and scaling them to strengthen useful task-specific knowledge. 

Through this dual strategy, SiGMA can mitigate performance degradation throughout the continual instruction tuning process by explicitly addressing negative interference. Since negative interference occurs in consolidating previously learned weights, merely preserving prior knowledge intact cannot tackle this interference.
As shown in Figure~\ref{fig:motiv}, the proposed SiGMA achieves promising improvements in mitigating performance degradation due to the negative interference while effectively preserving previously acquired representation knowledge. 

Finally, our contributions are summarized as follows: 
\begin{itemize}
    \item We propose SiGMA, a simple yet effective sign-decoupling framework that aims to mitigate negative interference in continual instruction tuning. To the best of our knowledge, it is the first work to introduce the sign-based weight decoupling idea in continual instruction tuning. 
    
    \item During training, the sign-guided adaptive tuning reduces knowledge collisions by adapting current LoRA weights to the general subspace, acquiring novel representations with minimal drift from prior knowledge.
    
    \item At inference, the sign-guided merging selectively amplifies crucial parameters in the specific subspace to retain its useful knowledge while reducing the impact of negative interference.
    
    \item SiGMA achieves state-of-the-art results on the UCIT (Unseen Continual Instruction Tuning) \cite{guo-etal-2025-hide} and DCL (Domain Continual Learning)~\cite{zhao2025mllm} benchmarks, and we further validate its effectiveness through extensive ablation studies and in-depth analysis experiments.
\end{itemize}


\section{Related Work}
\subsubsection{Multimodal Large Language Models.}
Instruction tuning has been crucial in aligning large language models~\cite{touvron2023llama,Jiang2023Mistral7} with human intention across diverse contexts. Recent visual instruction tuning~\cite{liu2023visual,liu2024improved} has extended this capability to multimodal large language models (MLLMs)~\cite{dai2023instructblip,Qwen-VL,zhu2023minigpt}, by integrating visual and text inputs for vision-language reasoning. However, most MLLMs are trained in an offline and static manner. This prevents MLLMs from adapting to the non-stationary nature of real-world tasks and evolving user requirements. In this work, we study multimodal continual instruction tuning, aiming to continuously improve the instruction following ability of MLLMs without severe forgetting that arises in sequential adaptation.

\subsubsection{Multimodal Continual Instruction Tuning.} 
Multimodal Continual Instruction Tuning (MCIT) requires MLLMs to continually follow instructions across successive tasks while preserving previously learned instruction-following abilities.
Prior MCIT methods primarily rely on Mixture-of-Experts~\cite{jacobs1991adaptive,shazeer2017outrageously} or expansion–merge strategies to adapt to new tasks while preserving earlier ones: CoIN~\cite{chen2024coin} introduces an MCIT benchmark of eight datasets and proposes MoE-LoRA to mitigate catastrophic forgetting. CL-MoE~\cite{huai2025cl} employs a dual-momentum MoE to route global and local experts. O-LoRA~\cite{wang2023orthogonal} leverages orthogonal subspace learning with orthogonal loss. HiDE-LLaVA~\cite{guo-etal-2025-hide} adopts task-specific expansion with task-general fusion. 

While existing approaches can mitigate severe forgetting, they overlook negative interference during inference-time consolidation, where newly learned updates can overwrite useful prior knowledge. 
The proposed SiGMA aims to alleviate the negative interference via sign-guided merging at inference and mitigate severe parameter drift through sign-guided adaptive tuning during training.


\begin{figure*}[t]
  \includegraphics[width=\linewidth]{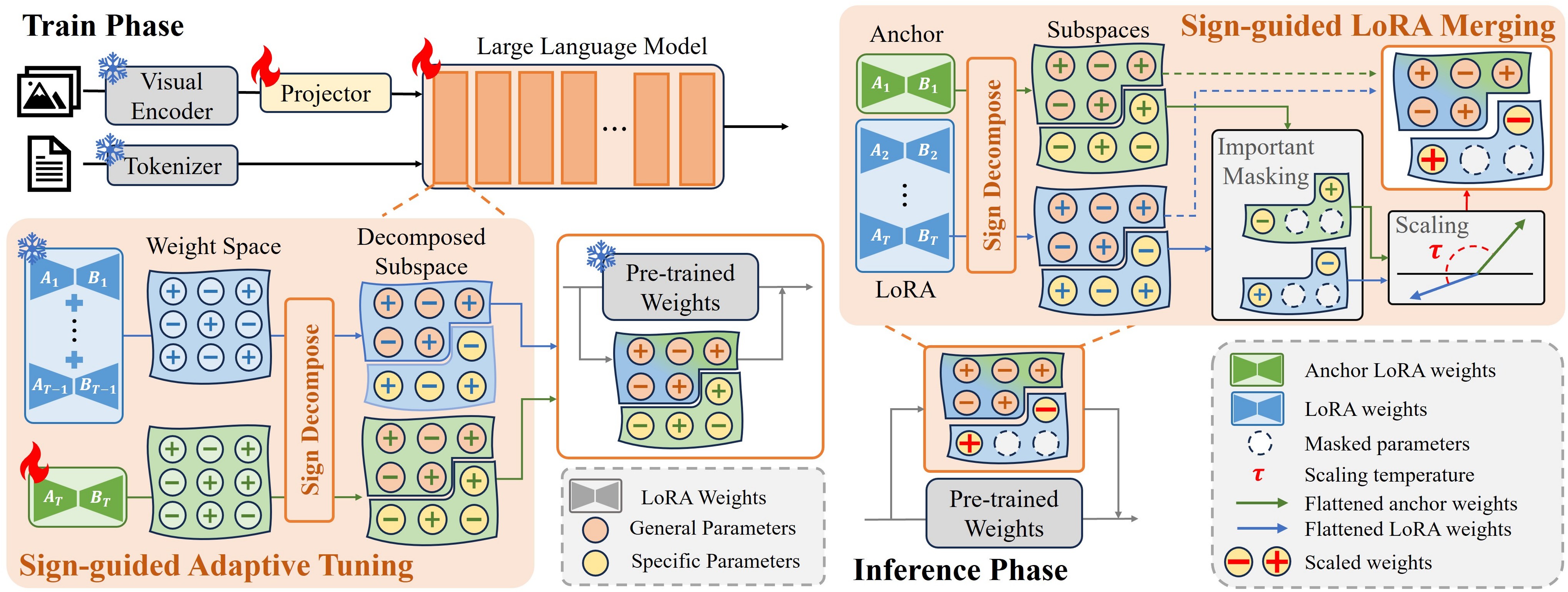}
  \caption {Overview of SiGMA. During training, sign-guided adaptive tuning (left) decomposes prior weights into general and task-specific subspaces to convey transferable knowledge. During inference, sign-guided LoRA merging (right) consolidates weights by selecting and scaling salient task-specific parameters, preserving crucial knowledge.}
  \label{fig:main}
\end{figure*}

\section{Methodology: SiGMA}
We first formulate the multimodal continual instruction tuning (MCIT) problem and present the base LoRA pipeline (\S~\ref{sec:3.1}).
We then analyze negative interference, which can be observed in the form of sign-conflicted parameters (\S~\ref{sec:3.2}).
Motivated by this, we introduce sign-based weight decoupling, the core mechanism of SiGMA (\S~\ref{sec:3.3}). 
Building on this, we detail sign-guided adaptive tuning, which leverages previously learned weights to efficiently adapt the model to the current task (\S~\ref{sec:3.4}). Finally, we present our sign-guided LoRA merging strategy, which selectively masks and amplifies important parameters to retain task-specific knowledge while minimizing negative interference (\S~\ref{sec:3.5}). Figure~\ref{fig:main} shows an overall framework of the proposed SiGMA.

\subsection{Preliminary}
\label{sec:3.1}
MCIT aims to learn from a continuous data stream while mitigating catastrophic forgetting. We consider a pre-trained MLLM parameterized by $\theta$ to be learned from a sequence of $T$ tasks. Each task $t \in \{1,\dots,T\}$ is defined by a dataset $D_t=\{(x_{vis}^{t,i}, x_{ins}^{t,i}, x_{ans}^{t,i})\}_{i=1}^{N_t}$, where $N_t$ denotes the number of image-text paired samples and $x_{vis}$, $x_{ins}$, and $x_{ans}$ denote the input image, instruction, and answer tokens, respectively.
Given an image–text input pair with an target answer of length $L$, the model is optimized to predict the next token in an autoregressive manner using the standard objective function:
\begin{equation}
  \label{eq:mllm_loss}
  \mathcal{L}^{t} = -\sum_{l=1}^{L}{\log p(x_{ans}^l \mid x_{vis}^t, x_{ins}^t, x_{ans}^{<l}; \theta)}.
\end{equation}
When learning task $t$, the goal of MCIT is to acquire novel knowledge from $\mathcal{D}_t$ while minimizing the forgetting of previous knowledge from $\mathcal{D}_{1:t-1}$. However, naive fine-tuning suffers from severe catastrophic forgetting, which disrupts not only the knowledge acquired from prior tasks but also the pre-trained foundational capabilities. Therefore, it is crucial to retain this foundational knowledge with the task-specific knowledge acquired during MCIT.

To achieve this, we utilize Parameter-Efficient Fine-Tuning (PEFT), which adapts large models to downstream tasks with minimal parameters. In the context of MCIT, Low-Rank Adaptation (LoRA)~\cite{hu2022lora} is widely adopted due to its effectiveness. 
Given a frozen pre-trained weight matrix $W \in \mathbb{R}^{d_{out}\times d_{in}}$, LoRA models the learnable weight $\Delta W=B\times A$, using two low-rank matrices $A\in\mathbb{R}^{r \times d_{in}}$ and $B\in\mathbb{R}^{d_{out} \times r}$, where the low-rank $r \ll \min(d_{out}, d_{in})$. The modified weight matrix $W'$ during training is formulated as
\begin{equation}
    W'= W+\Delta W = W+BA.
\end{equation}
By optimizing only $A$ and $B$, LoRA achieves efficient adaptation while keeping W frozen. In this work, we adopt an expansion-merge strategy as a base model. We train task-specific LoRA modules $(A_t,B_t)$ for each task $t$. At inference, we consolidate these parameters into unified weights $A'=\sum_{i=1}^t{A_i}$ and $B'=\sum_{i=1}^t{B_i}$ to enable comprehensive inference across all trained tasks.

\subsection{Negative Interference in Continual Instruction Tuning}
\label{sec:3.2}

Existing MCIT methods widely employ Mixture-of-Experts (MoE) or expansion-merge strategies to mitigate catastrophic forgetting. While effective in preserving learned knowledge, both approaches suffer from negative interference, where newly acquired knowledge conflicts with prior one. The negative interference harms the model's knowledge representation, as the weight consolidation can overwrite and distort preserved knowledge. 
Therefore, addressing negative interference is as crucial as mitigating catastrophic forgetting in MCIT.

Our research starts from the observation that the negative interference is highly correlated with the parameter signs. Given previously learned weights $W_{old}$ and newly learned weights $W_{new}$, we partition the parameter space into a sign-aligned subspace $R_{alig}$ and a sign-conflicted subspace $R_{conf}$.
When consolidating weights, we assume that the effects differ significantly across subspaces. In the sign-aligned subspace, weights share the same sign, so their magnitudes tend to accumulate, potentially reinforcing the learned knowledge. In contrast, in the sign-conflicted subspace, weights have opposing signs, which can cause cancellation during merging, reducing the effective magnitude and degrading the corresponding representation knowledge.
This is formulated as follows:
\begin{align}
    R_{alig} = \{i|\text{sgn}(W^{(i)}_{old})&=\text{sgn}(W^{(i)}_{new})\}, \;
    R_{conf} = \{j|\text{sgn}(W^{(i)}_{old}) \neq \text{sgn}(W^{(j)}_{new})\}, \\
    |W^{\mathcal{R}_{alig}}_{old} + W^{\mathcal{R}_{alig}}_{new}| &= |W^{\mathcal{R}_{alig}}_{old}| + |W^{\mathcal{R}_{alig}}_{new}| \quad (\text{Positive Interference}), \\
    |W^{\mathcal{R}_{conf}}_{old} + W^{\mathcal{R}_{conf}}_{new}| &\ll |W^{\mathcal{R}_{conf}}_{old}| + |W^{\mathcal{R}_{conf}}_{new}| \quad (\text{Negative Interference}). \label{eq:ni}
\end{align}
The negative interference due to opposing signs leads to a diminishing magnitude of the consolidated weights. This cancellation during weight consolidation may cause severe performance degradation despite the intention of consolidation to harmonize prior knowledge. 

\begin{figure}[t]
    \centering
    \includegraphics[width=0.49\linewidth]{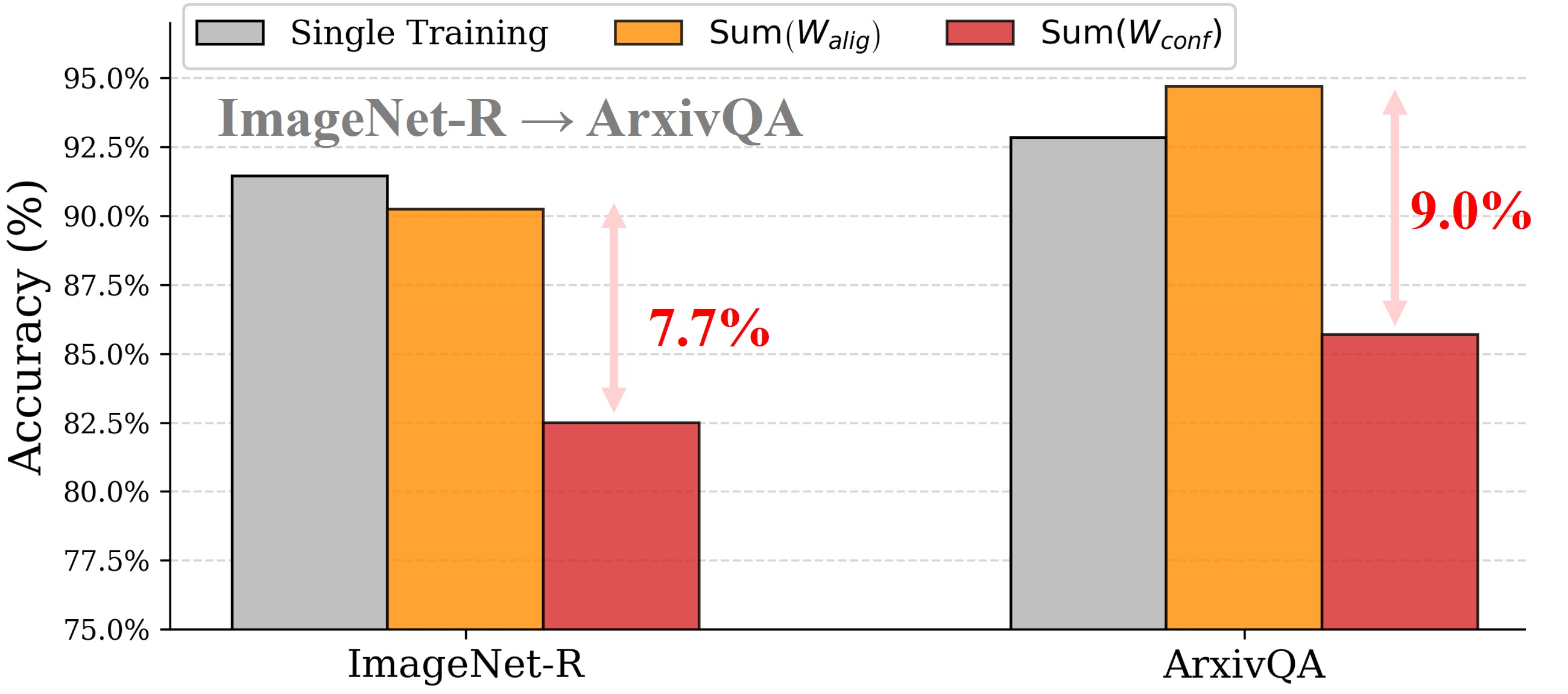}
    \hfill
    \includegraphics[width=0.49\linewidth]{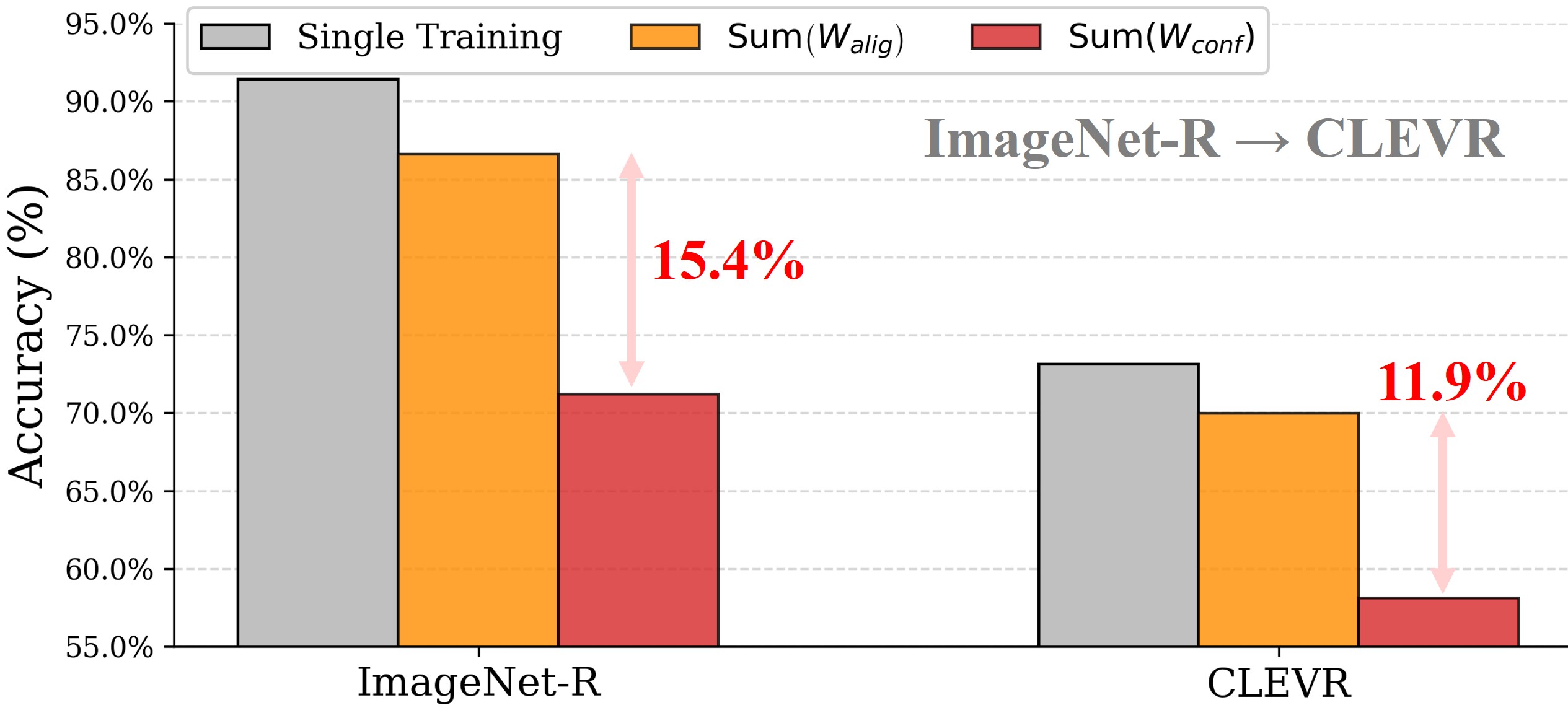}
    \caption{
    Impact of sign-based subspaces on continual learning performance. We train separate LoRA modules for ImageNet-R, ArxivQA, and CLEVR to preserve task-specific knowledge, and then merge them at inference. $W_{alig}$ represents merging only sign-aligned weights, while $W_{conf}$ represents merging only sign-conflicted weights.
    }
    \label{fig:NI_motiv}
\end{figure}

For empirical validation, we conduct experiments to analyze the effects of sign-aligned and sign-conflicted subspaces. We employ ``Single Training'' (\ie, independent training on each task) as an upper-bound baseline, representing the optimal performance for a specific task. As shown in~\Cref{fig:NI_motiv}, merging the sign-aligned subspace $W_{alig}$ preserves performance close to the baseline and, for ArxivQA, even surpasses the baseline, showing the aligned subspace tends to contribute to positive interference. In contrast, merging the sign-conflicted subspace $W_{conf}$ causes severe degradation. This hints that opposing signs tend to cause negative interference, where weight cancellation may lead imparing of prior knowledge and pose the weights in a suboptimal state, compromising both old task (\eg, ImageNet-R) and new tasks (\eg, ArxivQA and CLEVR). 
We provide diverse case studies in the supplementary material that further demonstrate the observed behaviors of sign-aligned and sign-conflicted subspaces.

\subsection{Sign-based Weight Decoupling}
\label{sec:3.3}

To address the negative interference in MCIT based on the observation in \S~\ref{sec:3.2}, we propose sign-based weight decoupling, which partitions weights into sign-aligned and sign-conflicted subspaces. 
Using the proposed decomposition, we not only leverage the sign-aligned subspace during the training to transfer prior positive knowledge, but also handle the sign-conflicted subspace in inference to mitigate negative interference. To the best of our knowledge, this is the first sign-based decoupling approach explicitly designed to mitigate negative interference.
Given the weights $W_{old}$ trained up to the previous task and the weights $W_{cur}$ for the current task, we compute binary general mask $M^g$ for the aligned subspace and the specific mask $M^s$ for the conflicting subspace as follows:
\begin{align}
    M^g = \mathbb{I}\left(\text{sgn}\left(W_{old}\right)=\text{sgn}\left(W_{new}\right)\right), \ \ 
    M^s = \mathbb{I}\left(\text{sgn}\left(W_{old}\right) \neq \text{sgn}\left(W_{new}\right)\right), \label{eq:mask_s}
\end{align}
where $\mathbb{I}(\cdot)$ and $\text{sgn}(\cdot)$ denote indicator function and sign function, respectively. 
We decompose $W_{old}$ into a general subspace $W_{old}^{g}=M^{g}\odot W_{old}$ and a specific subspace $W_{old}^{s}=M^{s}\odot W_{old}$, where $\odot$ denotes the element-wise product. The $M^{g}$ and $M^{s}$ masks form a partition over the parameters (\textit{i.e.}, $M^{g}\odot M^{s}=0$ and $M^{g}+M^{s}=\mathbf{1}$), ensuring that the sum of the general and specific subspaces exactly reconstructs the original old weights $W_{old}$.

Intuitively, $W_{old}^{g}$ contains the subset of parameters where the signs of $W_{old}$ and $W_{new}$ are aligned. Since this alignment indicates that the optimization landscapes of the previous and current tasks are consistent, $W_{old}^g$ represents task-invariant knowledge that can serve as a stable basis for learning the new task. In contrast, $W_{old}^s$ consists of parameters with opposing signs, indicating task-specific knowledge that is crucial for the previous tasks but conflicts with the information of current task.


\subsection{Sign-Guided Adapative Tuning}
\label{sec:3.4}
Building upon the sign-based weight decoupling strategy, we propose sign-guided adaptive tuning. This approach facilitates the learning of the current task $t$ by effectively leveraging transferable knowledge from previous tasks while filtering out conflicting information.

During the training of task $t$, we first construct a consolidated representaion of prior knowledge. We aggregate the low-rank matrices $A$ and $B$ separately across all $t-1$ previous tasks:
\begin{align}
    \Delta W_{prev} = \left(\sum_{i=1}^{t-1} B_i\right) \left(\sum_{i=1}^{t-1} A_i\right) 
    = \sum_{i=1}^{t-1} B_i A_i + \sum_{i \neq j} B_i A_j.
    \label{eq:cross-task}
\end{align}
By combining weights from different tasks, we explore a broader latent subspace that captures transferability across tasks, rather than limiting the model to the disjoint union of previous task spaces (\textit{i.e.}, applying only first term in Eq.~\ref{eq:cross-task}). Then, we dynamically decompose $\Delta W_{prev}$ based on the evolving current LoRA weights, $\Delta W_{cur}=B_tA_t$, to build a general subspace $\Delta W_{prev}^g$. We strictly utilize the sign-aligned general subspace to transfer knowledge to the current task. The modified forwarding process during training is as follows:
\begin{align}
    &W' = W_{\theta} + \Delta W_{prev}^g + \Delta W_{cur}, \mbox{ where } \\
    &\Delta W_{prev}^g =M^g \odot \Delta W_{prev}, \ \
    M^g =\mathbb{I}\left(\text{sgn}\left(\Delta W_{prev}\right)=\text{sgn}\left(\Delta W_{cur}\right)\right).
\end{align}
$W_{\theta}$ denotes the frozen pre-trained weights. The proposed sign-guided adaptive tuning has the property of dynamic adaptation. Although the aggregated prior weight $\Delta W_{prev}$ remains static, the extracted general subspace $\Delta W_{prev}^g$ updates throughout the training process. As the current weight $\Delta W_{cur}$ updates, the sign alignment shifts, allowing the model to adaptively retrieve different parameters for the prior knowledge that align with the current optimization direction. 

In summary, the proposed sign-guided adaptive tuning provides two key benefits. First, it facilitates the exploration of new knowledge by using the transferable parameters in $\Delta W_{prev}^g$ as a basis. Second, it imposes a soft regularization constraint. By ensuring alignment with the general subspace of prior knowledge, new task parameters do not deviate drastically from the previously learned weight space. Consequently, this alignment minimizes the conflict between $\Delta W_{cur}$ and $\Delta W_{prev}$, reducing the risk of negative interference during final consolidation.




\subsection{Sign-Guided LoRA Merging}
\label{sec:3.5}
While the proposed sign-guided adaptive tuning effectively aligns transferable knowledge during training, the remained task-specific subspaces tend to suffer from negative interference. Since these subspaces contain unique, task-dependent knowledge defined by divergent optimization directions, simply aggregating them leads to severe negative interference, where conflicting parameters disrupt each other. To address this at the inference stage, we propose sign-guided LoRA merging, which selectively strengthens salient task-specific parameters while filtering out noise-like parameters, ensuring that the unique characteristics of each task are preserved within the unified model.

After training up to task $t$, weight merging is needed to infer the model to represent the overall learned knowledge. To partition between general and specific subspaces consistently across all tasks, we use the LoRA weights from the first task $\Delta W_1$ as the anchor weights. $\Delta W_1$  should provide the stable basis for defining the parameter space, since it serves as the adaptation of the pre-trained model and is utilized through all subsequent training tasks in adaptive tuning. For every subsequent task $k \in \{2,...,t\}$, we decouple the weights $\Delta W_k$ based on the anchor $\Delta W_1$ into general subspace $\Delta W_k^g$, whose signs align with those of the anchor and specific subspace $\Delta W_k^s$, whose signs oppose those of the anchors.
The sign-guided decomposition by anchor weights can be formulated as
\begin{equation}
    \Delta W_k^g= M_k^g \odot \Delta W_k, \ \ 
    \Delta W_k^s= M_k^s \odot \Delta W_k,
\end{equation}
where $M_k^g$ and $M_k^s$ denote binary masks computed by~\cref{eq:mask_s}.
The general subspace $\Delta W_k^g$ captures transferable updates that can be safely aggregated, while the specific subspace $\Delta W_k^s$ contains opposite-sign updates and therefore must be handled carefully to avoid knowledge collisions.

To mitigate this collision, we propose a selecting and scaling strategy. We first filter $\Delta W_k^s$ to retain only the most significant parameters (\textit{e.g.}, those with large magnitudes), thereby removing small noise-like weights that may contribute to interference. Then, we amplify the selected parameters using a scaling factor computed from the cosine distance $d_k = \delta(\Delta W_1^s, \Delta W_k^s)$ between selected specific parameters from $\Delta W_1^s$ and those from $\Delta W_k^s$. Since the specific subspace has opposite signs to the anchor, the distance is empirically greater than 1 (\textit{i.e.}, $d_k \ge 1$). Thus, the sign-guided LoRA merging is formulated as
\begin{align}
    \Delta W_k^s = \Delta W_k \odot \hat{M_{k}^{s}}, \mbox{ where }
    \hat{M_{k}^{s}}=M_k^s \wedge \mathbb{I}(|\Delta W_k|>\lambda|\Delta W_1|),
\end{align}
$\lambda$ and $\hat{M_k^s}$ denote a temperature factor for anchor weights and the selected binary mask for a specific subspace. For inference, we unify all trained LoRA weights with the pre-trained weight $W_{\theta}$. The final unified weight merging becomes
\begin{equation}
    W'=W_{\theta}+\Delta W_1+\sum_{k=2}^t{(\Delta W_k^g+d_k\Delta W_k^s}).
\end{equation}
In summary, we identify salient parameters within the task-specific weights and scale them based on the degree of their specificity relative to the anchor weights. By suppressing the integration of insignificant weights while amplifying crucial task-specific weights, the proposed merging strategy effectively minimizes interference during the weight consolidation process.
We summarize the overall algorithm of the SiGMA in the Appendix. 


\newcolumntype{g}{>{\columncolor{gray!15}}c}
\newcolumntype{y}{>{\columncolor{yellow!15}}c}


\begin{table}[t]
\resizebox{\linewidth}{!}{%
\setlength{\tabcolsep}{2pt} 
\renewcommand{\arraystretch}{1.4} 
\begin{tabular}{cccccccgy}
\specialrule{2pt}{1pt}{1pt}
UCIT & ImageNet & Arxiv & VizWiz & IconQA & CLEVR & Flicker30k & Avg. All & Forgets \\
\hline
Zero-shot & 16.20 & 53.73 & 38.34 & 19.20 & 20.67 & 41.84 & 31.66 & - \\
\hdashline
LoRA-FT {[}ICLR'22{]} & 61.90 & 78.20 & 44.52 & 49.73 & 52.70 & \underline{57.53} & 57.43 & -17.02 \\
O-LoRA {[}EMNLP’23{]} & 72.73 & 77.77 & 43.87 & 45.83 & \underline{55.33} & 57.27 & 58.80 & -13.70 \\
MoE LoRA {[}NeurIPS’24{]} & 68.73 & 77.83 & 44.41 & 49.80 & 50.87 & 57.37 & 58.17 & -15.24 \\
CL-MoE {[}CVPR’25{]} & 68.00 & 77.50 & 43.92 & \textbf{52.37} & 53.17 & \textbf{57.80} & 58.79 & -15.29 \\
HiDe LLaVA {[}ACL’25{]} & \textbf{84.70} & 90.53 & \underline{49.98} & 49.43 & 45.07 & 52.30 & 62.00 & \underline{-5.58} \\
DISCO {[}ICCV’25{]} & 80.83 & \underline{92.07} & 46.24 & 50.70 & 50.97 & 56.66 & \underline{62.91} & -6.27 \\
\hdashline
\textbf{SiGMA (Ours)} & \underline{83.40} & \textbf{94.07} & \textbf{54.83} & \underline{51.03} & \textbf{57.83} & 55.87 & \textbf{66.17} & \textbf{-4.81} \\ \specialrule{2pt}{1pt}{1pt}
\end{tabular}%
}
\vfill
\resizebox{\linewidth}{!}{%
\setlength{\tabcolsep}{6pt} 
\renewcommand{\arraystretch}{1.4} 
\begin{tabular}{ccccccgy}
\specialrule{2pt}{1pt}{1pt}
DCL & RS & MED & AD & Sci & Fin & Avg. All & Forgets \\
\hline
Zero-shot & 32.29 & 28.28 & 15.59 & 35.55 & 62.56 & 34.85 & - \\
\hdashline
LoRA-FT {[}ICLR'22{]} & 69.65 & 41.59 & 25.43 & 40.88 & 87.45 & 53.00 & -14.97 \\
O-LoRA {[}EMNLP’23{]} & 76.04 & 44.71 & 36.60 & 46.35 & \underline{90.19} & 58.78 & -9.75 \\
MoE LoRA {[}NeurIPS’24{]} & 77.16 & 47.76 & 30.77 & 41.34 & 89.50 & 57.31 & -12.19 \\
CL-MoE {[}CVPR’25{]} & 72.91 & 49.67 & 36.00 & 42.93 & \textbf{90.22} & 58.35 & -11.56 \\
HiDe LLaVA {[}ACL’25{]} & \textbf{78.84} & \underline{51.22} & \underline{38.09} & \underline{46.68} & 85.11 & 59.99 & \underline{-3.86} \\
DISCO {[}ICCV’25{]} & 75.60 & 44.07 & \textbf{51.68} & 45.40 & 85.36 & \underline{60.42} & -5.44 \\
\hdashline
\textbf{SiGMA (Ours)} & \underline{78.20} & \textbf{59.75} & 35.36 & \textbf{50.73} & 87.45 & \textbf{62.30} & \textbf{-2.14} \\ \specialrule{2pt}{1pt}{1pt}
\end{tabular}%
}
\caption{Comparison with recent methods on the UCIT (top) and DCL (bottom) benchmark in terms of Last, Avg. All, and Forgetting. The best and second-best results are highlighted in bold and underline, respectively.}
\label{tab:main_result}
\end{table}

\section{Experiments}
We demonstrate that SiGMA effectively mitigates negative interference and subsequently improves the performance of MCIT across various benchmarks.

\subsection{Experimental Settings}
\subsubsection{MCIT Benchmarks.} 
We evaluate SiGMA on two behcnmarks: (1) UCIT (Unseen Continual Instruction Tuning) \cite{guo-etal-2025-hide} includes six instruction-tuning datasets including ImageNet-R~\cite{hendrycks2021many}, ArixivQA~\cite{li2024multimodal}, Vizwiz~\cite{gurari2018vizwiz}, IconQA~\cite{lu2021iconqa}, CLEVR-math~\cite{lindstrom2022clevr}, and Flicker30k~\cite{plummer2015flickr30k}. They are chosen for LLaVA’s weak zero-shot performance and filtered to avoid overlap with LLaVA’s pre-training data, minimizing information leakage. (2) DCL (Domain Continual Learning)~\cite{zhao2025mllm} can test the model trained sequentially across multiple domains, including remote sensing~\cite{lobry2020rsvqa}, medical~\cite{he2020pathvqa}, autonomous driving~\cite{sima2024drivelm}, science~\cite{kembhavi2016diagram,guo2025sciverse,chang2022mapqa,kembhavi2017you}, and finance~\cite{wang2023finvis,zhao2025mllm}. The severe domain shifts in the DCL benchmark often lead to strong cross-domain interference and significant forgetting.

\subsubsection{Evaluation Metrics.}
Following standard continual learning evaluation protocols~\cite{wang2022learning,guo-etal-2025-hide}, we report \textit{Last}, \textit{Average}, and \textit{Forgets} to assess the performance. (1) \textit{Last} is the accuracy on each previously seen task measured after training on the final task.
(2) \textit{Average} is reported in two forms; \textit{Avg. All} is the mean of the last accuracies across all tasks, and \textit{Avg. Each} is the average accuracy evaluated immediately after the last task.
(3) \textit{Forgets} is the average drop in accuracy for each task, defined as the difference between its accuracy immediately after training and its final accuracy after the last task. In summary, \textit{Avg. All}, \textit{Avg. Each}, and \textit{Forgets} respectively reflect overall knowledge retention, the ability to learn new tasks, and the extent of catastrophic forgetting in continual learning.

\subsubsection{Implementation details.}
We use LLaVA-v1.5-7B~\cite{liu2024improved} as the backbone model in all experiments. Following the LoRA fine-tuning setup from LLaVA, we insert LoRAs into the linear layers of the language model with a low-rank set to 16. We apply the proposed sign-guided adaptive tuning to all layers except the final output layers for next-token generation. For sign-guided LoRA merging, we set the temperature $\lambda=0.8$. We set a batch size of 32 and 8 for all methods on UCIT and DCL benchmarks. We run all experiments on RTX A6000 GPUs.

\subsection{Main Results}
We compare SiGMA with state-of-the-art LoRA-based continual instruction tuning baselines such as LoRA-FT~\cite{hu2022lora}, O-LoRA~\cite{wang2023orthogonal}, MoE LoRA~\cite{chen2024coin}, CL-MoE~\cite{huai2025cl}, Hide-LLaVA~\cite{guo-etal-2025-hide}, and DISCO~\cite{guo2025federated} on UCIT and DCL benchmarks. As shown in Table~\ref{tab:main_result}, SiGMA achieves the best Avg. All and Forgets on UCIT, indicating strong robustness to severe forgetting and negative interference. While HiDe-LLaVA and DISCO reduce forgetting via task-specific expansion and task-general fusion with auxiliary text encoders~\cite{radford2021learning}, SiGMA reports state-of-the-art performance without auxiliary models by mitigating interference at inference time. Notably, although all baselines score below 50.00\% on VizWiz due to its difficulty and severe forgetting, SiGMA improves over the second-best method by +4.85\%.

On the DCL benchmark, where each task requires domain-specific knowledge and the domain shifts are severe, prior methods suffer from severe forgetting, as shown in Table~\ref{tab:main_result}. In contrast, SiGMA achieves the best performance, outperforming the second-best method by +1.88\% in Avg. All and +1.72\% in Forgets. These results show that SiGMA preserves prior knowledge more effectively and remains robust under severe domain shift.

\begin{table}[t]
\begin{subtable}[t]{0.48\columnwidth} 
\centering
\resizebox{\columnwidth}{!}{%
\setlength{\tabcolsep}{2pt} 
\renewcommand{\arraystretch}{1.4} 
\begin{tabular}{ccccc}
\specialrule{2pt}{1pt}{1pt}
\multicolumn{2}{c}{SiGMA} & \multicolumn{3}{c}{Metrics}    \\ \hline
S-Tuning     & S-Merge    & Avg. All & Avg. Each & Forgets \\ \hline
             &            & 62.42    & 67.41     & -5.98   \\ \hdashline
             & \checkmark & 63.35    & 68.57     & -6.26   \\
\checkmark   &            & 64.86    & 68.56     & \textbf{-4.44}   \\ \hdashline
\checkmark   & \checkmark & \textbf{66.17}    & \textbf{70.18}     & -4.81  \\ \specialrule{2pt}{1pt}{1pt}
\end{tabular}%
}
\caption{
Ablation studies on SiGMA. 
}
\label{tab:ablation}
\end{subtable}
\hfill
\begin{subtable}[t]{0.48\columnwidth} 
\centering
\resizebox{\columnwidth}{!}{%
\setlength{\tabcolsep}{1.2pt} 
\renewcommand{\arraystretch}{1.4} 
\begin{tabular}{ccccc}
\specialrule{2pt}{1pt}{1pt}
\multicolumn{2}{c}{S-Tuning} & \multicolumn{3}{c}{Metrics} \\ \hline
LoRA $\Delta$ & SWD & Avg. All & Avg. Each & Forgets     \\ \hline
$\sum_{i}{B_iA_i}$ &   & 63.23 & 67.85 & -5.54 \\
$\sum_{i}{B_i}\sum_{i}{A_i}$ &   & 62.42 & 67.41 & -5.98 \\ \hdashline
$\sum_{i}{B_iA_i}$ & \checkmark & 64.15 & \textbf{69.06} & -5.89          \\
\textbf{$\mathbf{\sum_{i}{B_i}\sum_{i}{A_i}}$} & \checkmark & \textbf{64.86} & 68.56 & \textbf{-4.44} \\ \specialrule{2pt}{1pt}{1pt}
\end{tabular}%
}
\caption{
Analysis of S-Tuning. 
}
\label{tab:SAT_analysis_tuning}
\end{subtable}
\caption{Ablation studies on S-Tuning and S-Merge of SiGMA (left), with analysis of S-Tuning (right). We examine the contribution of each component of SiGMA and analyze the effects of weight consolidation and sign-based weight decoupling (SWD).}
\begin{subtable}[t]{0.48\columnwidth} 
\centering
\resizebox{\columnwidth}{!}{%
\setlength{\tabcolsep}{2pt} 
\renewcommand{\arraystretch}{1.3} 
\begin{tabular}{cccc}
\specialrule{2pt}{1pt}{1pt}
S-Tuning  & \multicolumn{3}{c}{Metrics} \\ \hline
w/o top-k layers                    & Avg. All       & Avg. Each      & Forgets     \\ \hline
k = 32                        & 63.35          & 68.57          & -6.26          \\ \hdashline
k = 16                        & 64.41          & 68.48          & -4.88          \\
k = 5                         & 65.55          & 68.32          & -5.38          \\
k = 3                         & 65.56          & 69.74          & -5.01          \\ \hdashline
\textbf{k=1} & \textbf{66.17} & \textbf{70.18} & \textbf{-4.81} \\ \hdashline
k = 0 (all layer)             & 64.27          & 68.49          & -5.07          \\ \specialrule{2pt}{1pt}{1pt}
\end{tabular}%
}
\caption{
Analysis of S-Tuning across the layer depth.
}
\label{tab:S_tuning_layer}
\end{subtable}
\hfill
\begin{subtable}[t]{0.48\columnwidth} 
\centering
\resizebox{\columnwidth}{!}{%
\setlength{\tabcolsep}{2pt} 
\renewcommand{\arraystretch}{1.2} 
\begin{tabular}{cccc}
\specialrule{2pt}{1pt}{1pt}
S-Merge & \multicolumn{3}{c}{Metrics}                            \\ \hline
Temp $\lambda$ (w/ $d_k$)                               & Avg. All       & Avg. Each      & Forgets              \\ \hline
$\lambda$ = 0. (w/o $\lambda$)                            & 54.32          & 70.69          & -19.65               \\ \hdashline
$\lambda$ = 0.6                & 65.15          & \textbf{70.56} & -6.49                \\
$\lambda$ = 0.7                                          & 65.88          & 70.40          & -5.43                \\ \hdashline
$\lambda$ = 0.8 (w/o $d_k$)     & 62.69          & 67.11          & -5.30                \\
$\bm{\lambda}$ \textbf{= 0.8}                & \textbf{66.17} & 70.18          & -4.81                \\ \hdashline
$\lambda$ = 0.9                & 65.99          & 69.96          & -4.76                \\
$\lambda$ = 1.0                                          & 65.89          & 69.81          & \textbf{-4.70} \\ \specialrule{2pt}{1pt}{1pt}
\end{tabular}%
}
\caption{
Analysis of sign-guided LoRA merging. 
}
\label{tab:S_merge_analysis_temp}
\end{subtable}
\caption{Analysis of S-Tuning and S-Merge. We analyze the effect of S-Tuning across layer depth (left) and the impact of the temperature factor $\lambda$ and amplification term $d_k$ in S-Merge (right) in terms of Average and Forgets.}
\end{table}

\subsection{Further Analyses}
To assess the effectiveness of SiGMA in more detail, we conduct additional experiments on the UCIT benchmarks. 
The proposed SiGMA has two main components: sign-guided adaptive tuning (coined as S-Tuning) and sign-guided LoRA merging (S-Merge). 
To gauge their individual contributions, we first perform ablations. 
As further analysis for S-Tuning, we test the effect of sign-based weight decoupling (SWD) and the layer position. For S-Merge, we ablate the temperature factor and evaluate the effects of amplification.

\subsubsection{Ablation Study.}
We report the performance of ablations in Table~\ref{tab:ablation}. As a baseline, we adopt an expansion–merge strategy that trains task-specific LoRA modules and merges all learned LoRAs at inference. 
Although this baseline can preserve prior weights to reduce severe forgetting, it remains susceptible to interference during consolidation. 
Applying S-Tuning or S-Merge alone still improves overall performance. However, S-Merge reports higher accuracy but slightly worse forgetting, since it does not explicitly enforce a discriminative separation between general and specific subspaces during training and thus may merge conflicting updates. S-Tuning consistently reduces forgetting by promoting more transferable updates that remain compatible when all task LoRAs are aggregated at inference. Combining S-Tuning and S-Merge achieves the best results, demonstrating their promising synergistic effect.

\subsubsection{Analysis on Sign-guided Adaptive Tuning (S-Tuning).}
Table~\ref{tab:SAT_analysis_tuning} analyzes the effects of sign-based weight decoupling (SWD) within S-Tuning and compares two consolidation strategies: merging the full LoRA updates $\Delta$ and merging the low-rank matrices $A$ and $B$. SWD consistently improves overall performance under both strategies, supporting its role in conveying transferable knowledge from prior tasks. Notably, consolidating $A$ and $B$ is more effective than consolidating full LoRA weights $\Delta$ when SWD is used. While $\Delta$-level consolidation simply combines task-specific knowledge, low-rank consolidation combines low-rank components across tasks, which better explores general latent subspaces. Combined with SWD, this provides a transferable subspace during training, reducing drift from previous tasks and improving adaptation to new tasks.

Furthermore, we investigate where to apply S-Tuning across layers. As shown in Table~\ref{tab:S_tuning_layer}, we  increase the number of S-tuned layers by applying S-Tuning to all layers except the top-k layers of the LLM. Here, $k=32$ means absence of S-Tuning, and $k=0$ means S-Tuning applied to all layers.
Overall performance increases as S-Tuning is applied to more layers, except when all layers are tuned. This performance drop is expected because the top-1 layer directly connects to the next-token prediction, injecting overly task-invariant knowledge can degrade discriminative decisions. Accordingly, excluding the top-1 layer shows strong Avg. All and Avg. Each scores and the lowest Forgets.

\begin{figure}[t]
  \includegraphics[width=0.5\columnwidth]{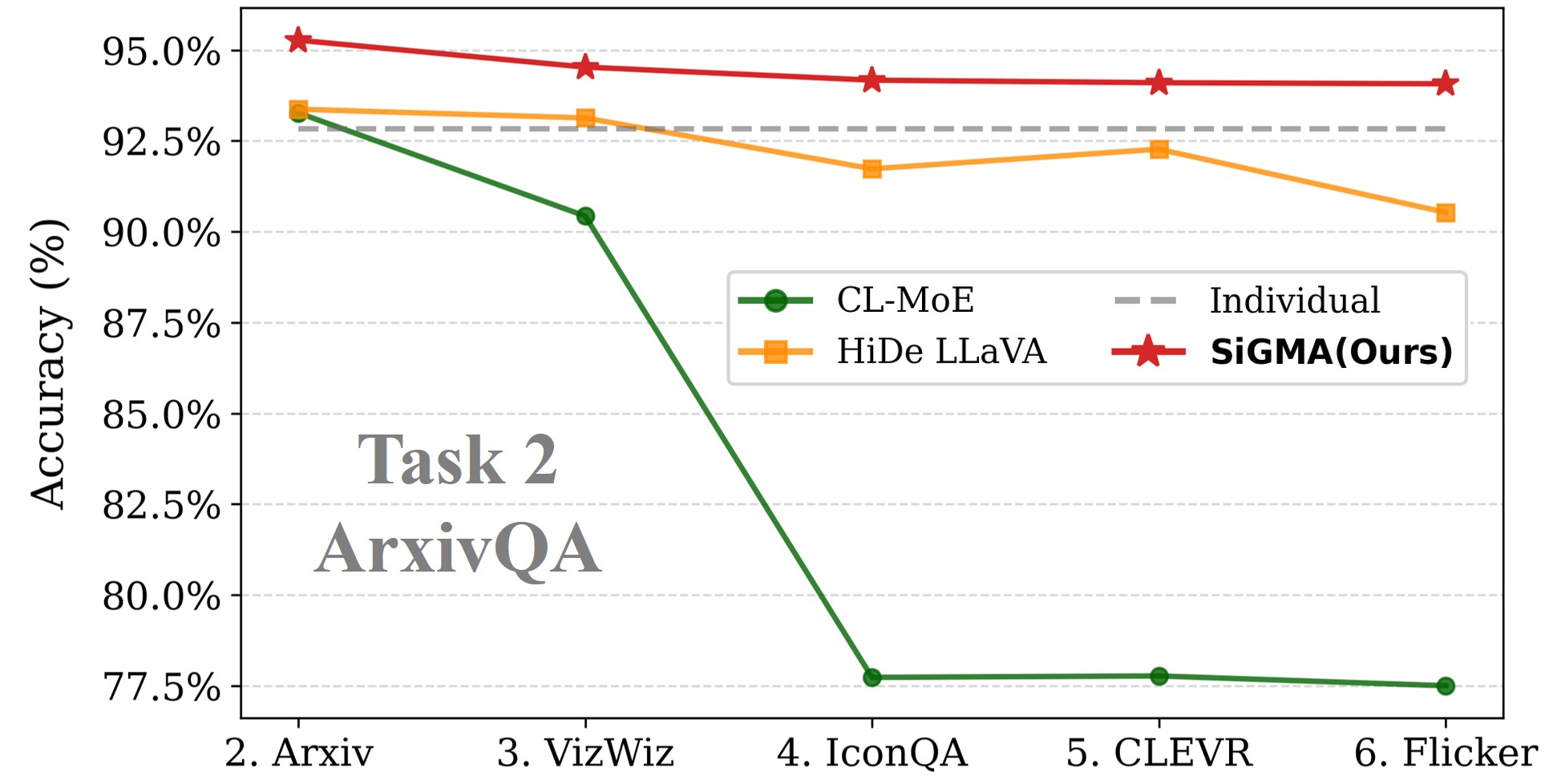}
  \hfill
  \includegraphics[width=0.5\columnwidth]{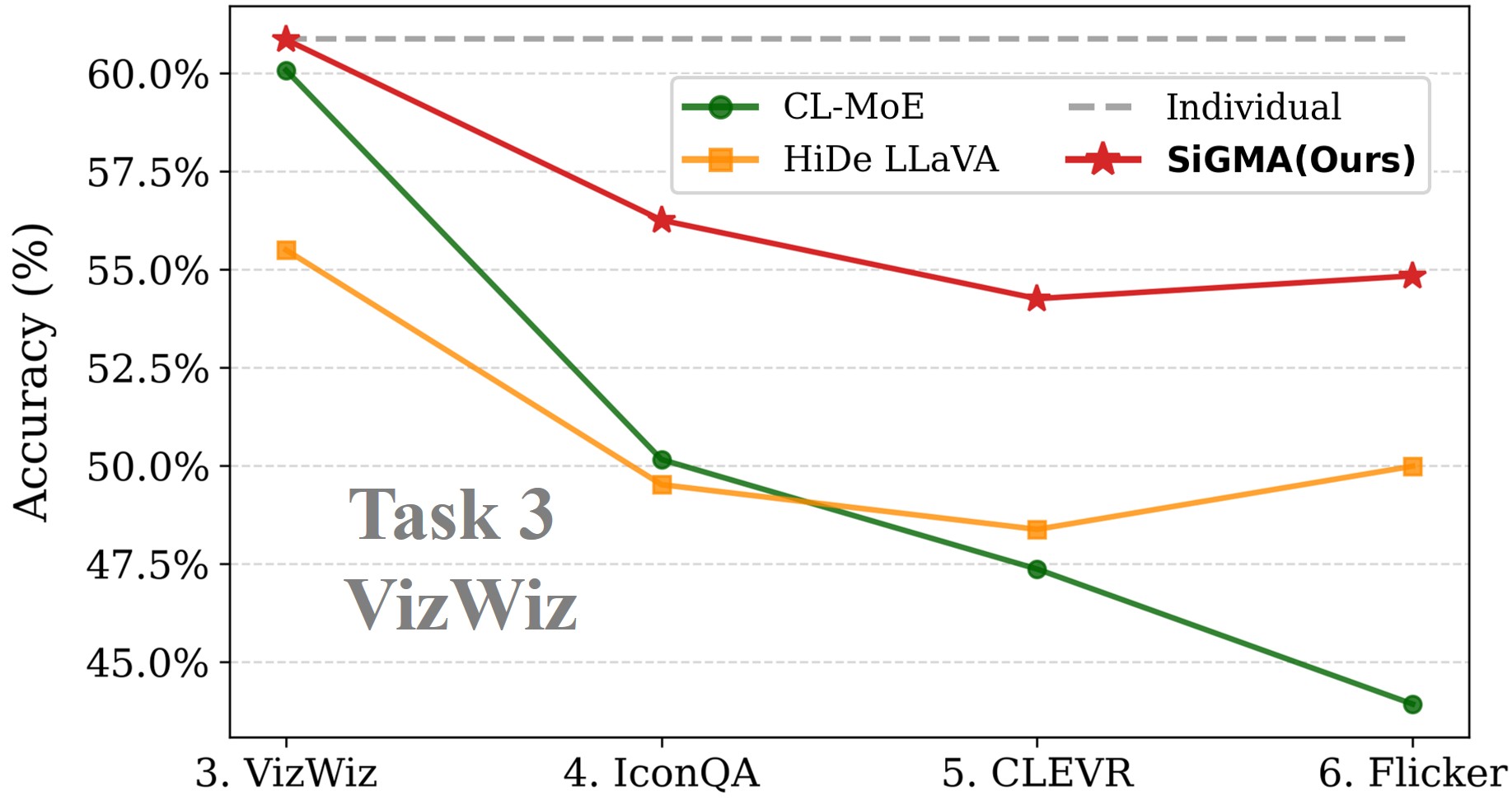}
  \caption{
  Performance variation under the UCIT benchmark. We report changes in single-task performance (\eg, ArxivQA and VizWiz) as the model sequentially learns new tasks. “Individual” denotes the performance achieved when the model is trained only on the corresponding task.
  }
  \label{fig:analysis_forgets}
\end{figure}

\subsubsection{Analysis on Sign-guided LoRA Merging (S-Merge).}
We first examine how the varied temperature factor $\lambda$ affects the selection of salient parameters. As shown in Table~\ref{tab:S_merge_analysis_temp}, increasing the temperature $\lambda$ improves overall performance up to 0.8. Although the best Avg. Each and Forgets report at slightly different temperatures, we set $\lambda=0.8$ because it best reflects overall knowledge retention. 
Using the optimal temperature, we then test the effects of weight amplification by cosine distance $d_k$.
Removing $d_k$ causes a clear degradation: Avg. All and Avg. Each drop by 3.48\% and 3.07\%, respectively, and Forgets increases by 0.49\%. These results demonstrate that amplifying selected parameters helps preserve their knowledge when consolidating weights at inference.

As $\lambda$ increases, we observe a clear trade-off; Avg. Each slightly declines, while Forgets improves significantly. Specifically, the degradation in Avg. Each is marginal (dropping from 70.69\% to 69.81\%), whereas the reduction in forgetting is substantial (improving from -19.65\% to -4.70). These results demonstrate that enforcing stricter parameter selection, which restricts the number of specific parameters merged during inference, sacrifices only a minimal amount of plasticity (performance on the current task) in exchange for a substantial gain in stability. 
Intuitively, S-Merge enables the weights to filter out less critical parameters, preventing negative interference while preserving prior knowledge.

\subsubsection{Analysis on Negative Interference.}
To validate mitigated negative interference, we track the performance of specific tasks (\eg, ArxivQA and VizWiz) as the model sequentially learns subsequent tasks. 
As shown in~\Cref{fig:analysis_forgets},  prior methods still suffer from severe degradation due to negative interference. Specifically, CL-MoE shows a drastic decline as new tasks are introduced, and HiDe LLaVA, despite preserving prior weights intact, reports sub-optimal performance. In contrast, SiGMA achieves outstanding performance in preserving prior knowledge. Notably, in ArxivQA, SiGMA consistently outperforms individual training throughout the entire learning process. This shows that SiGMA effectively mitigates negative interference, improving positive transferability. 

\subsubsection{Qualitative Results.}
\begin{figure}[t]
  \includegraphics[width=0.5\columnwidth]{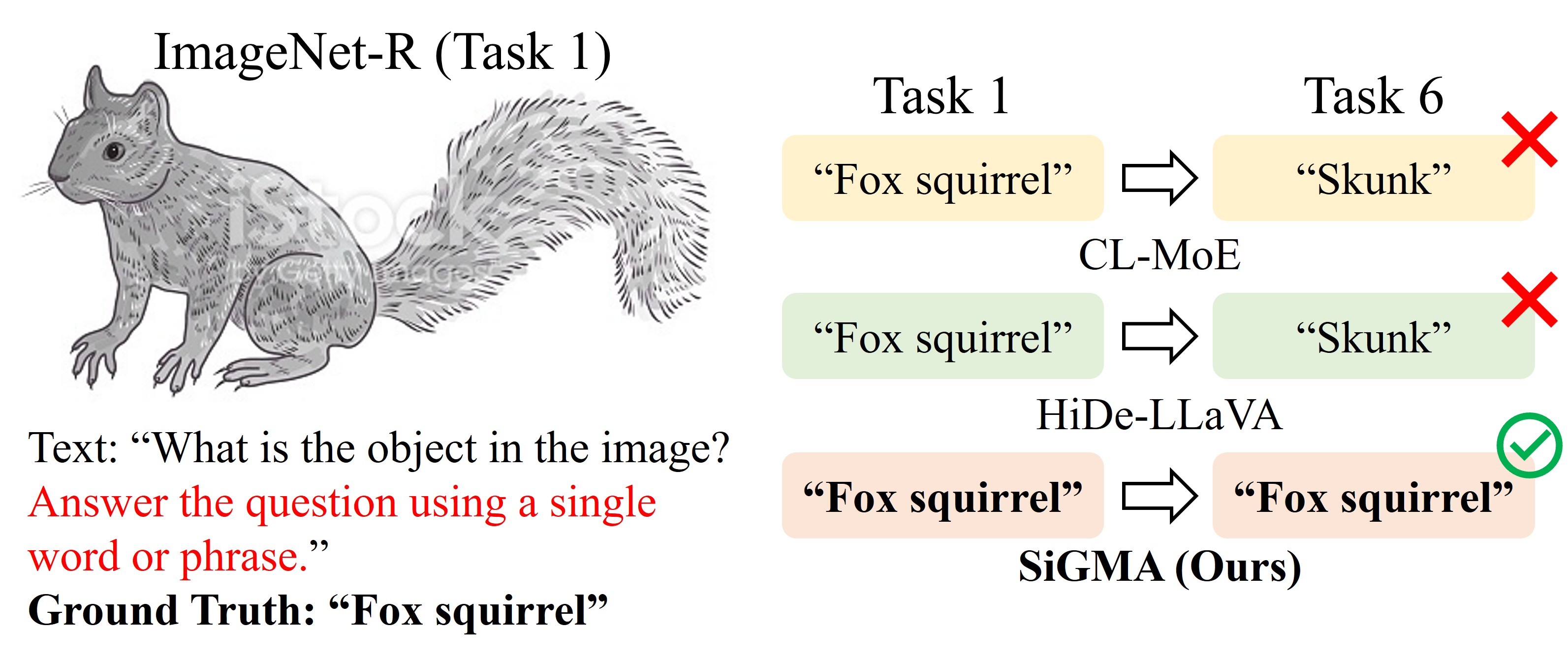}
  \hfill
  \includegraphics[width=0.5\columnwidth]{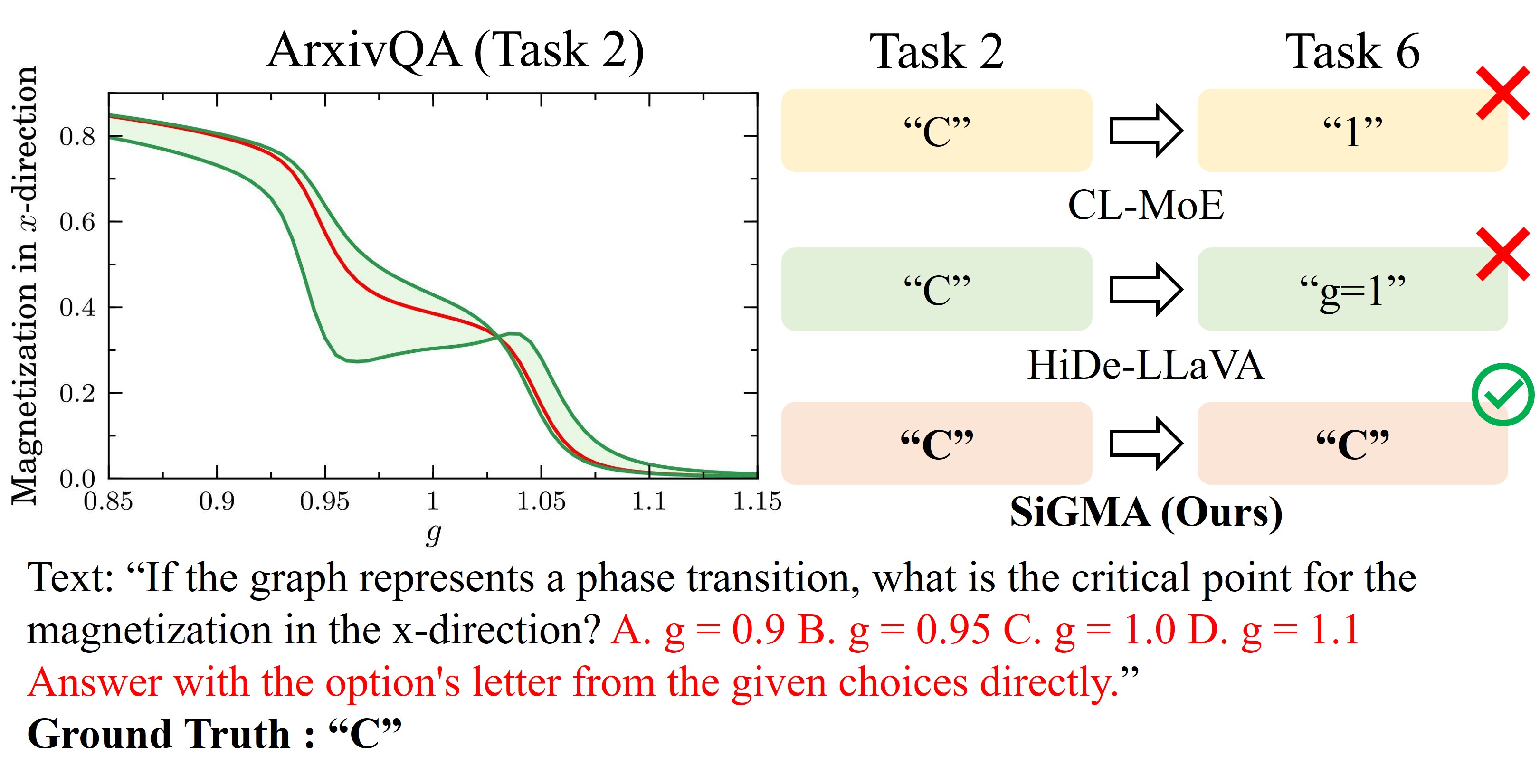}
  \caption{
  Qualitative examples from ImageNet-R (left) and ArxivQA (right) on the UCIT benchmark. Given the same image and instruction, we compare responses from CL-MoE, HiDe-LLaVA, and SiGMA, shown both after task-specific training and after sequential training on all tasks. Red text denotes the task-specific instruction format that the model must follow to generate a response.
  }
  \label{fig:quality_analysis}
\end{figure}

Figure~\ref{fig:quality_analysis} shows 
qualitative examples on the UCIT benchmark, comparing SiGMA with state-of-the-art methods: CL-MoE and HiDE-LLaVA.
All methods follow instructions and produce correct answers after training on the corresponding task. However, after sequentially training all tasks, CL-MoE and HiDE-LLaVA 
suffer from generating incorrect responses on ImageNet-R (\eg, misclassifying a ``Fox squirrel'' as a ``Skunk'') and failing instruction following (\eg, generating the full content of an answer option instead of simply producing the corresponding option letter).
In contrast, SiGMA maintains strong instruction-following ability and preserves correct task knowledge. 
By effectively mitigating negative interference, it ensures that the model reliably adapts to new, task-specific instructions without suffering from catastrophic forgetting.

\section{Conclusion}
We address negative interference, a critical yet often overlooked challenge in MCIT, where the consolidation of new weights disrupts previously acquired knowledge. Unlike prior methods that primarily focus on catastrophic forgetting, SiGMA explicitly aims to mitigate negative interference. 
Based on sign-based weight decoupling, our sign-guided adaptive tuning enables stable and transferable learning with minimal parameter drift, while sign-guided LoRA merging mitigates negative interference by selectively amplifying salient parameters. 

Despite its strong performance, SiGMA still has room for improvement. The general subspace may become increasingly sparse as the task sequence grows, reducing its effectiveness in long-term continual instruction tuning, for which future work will explore strategies to robustly maintain the general subspace. 
Furthermore, we plan to extend sign-based weight decoupling beyond LoRA to other Parameter-Efficient Fine-Tuning (PEFT) techniques, such as adapters and learnable prompts. Given their widespread adoption for efficient adaptation, we believe that extending sign-based interference across diverse PEFT architectures will shed light on broader insights into the continual learning paradigm.

\section*{Acknowledgements}
This work was supported by Samsung Electronics Co., Ltd (IO250418-12669-01) and Institute of Information \& communications Technology Planning \& Evaluation (IITP) grant funded by the Korea government (MSIT): (1) No.~RS-2026-25524173, Ultro-Long-Term Hierarchical Memory and Reasoning Architecture for Next-Generation Omnimodal Agents, (2) No.~RS-2019-II191082, SW StarLab, (3) No.~RS-2022-II220156, Fundamental research on continual meta-learning for quality enhancement of casual videos and their 3D metaverse transformation, and (4) No.~RS-2021-II211343, Artificial Intelligence Graduate School Program (Seoul National University).
Gunhee Kim is the corresponding author.






%
%
\bibliographystyle{splncs04}
\bibliography{main}

@inproceedings{chen2024internvl,
  title={Internvl: Scaling up vision foundation models and aligning for generic visual-linguistic tasks},
  author={Chen, Zhe and Wu, Jiannan and Wang, Wenhai and Su, Weijie and Chen, Guo and Xing, Sen and Zhong, Muyan and Zhang, Qinglong and Zhu, Xizhou and Lu, Lewei and others},
  booktitle={Proceedings of the IEEE/CVF conference on computer vision and pattern recognition},
  pages={24185--24198},
  year={2024}
}

@article{liu2023visual,
  title={Visual instruction tuning},
  author={Liu, Haotian and Li, Chunyuan and Wu, Qingyang and Lee, Yong Jae},
  journal={Advances in neural information processing systems},
  volume={36},
  pages={34892--34916},
  year={2023}
}

@article{dai2023instructblip,
  title={Instructblip: Towards general-purpose vision-language models with instruction tuning},
  author={Dai, Wenliang and Li, Junnan and Li, Dongxu and Tiong, Anthony and Zhao, Junqi and Wang, Weisheng and Li, Boyang and Fung, Pascale N and Hoi, Steven},
  journal={Advances in neural information processing systems},
  volume={36},
  pages={49250--49267},
  year={2023}
}

@article{achiam2023gpt,
  title={Gpt-4 technical report},
  author={Achiam, Josh and Adler, Steven and Agarwal, Sandhini and Ahmad, Lama and Akkaya, Ilge and Aleman, Florencia Leoni and Almeida, Diogo and Altenschmidt, Janko and Altman, Sam and Anadkat, Shyamal and others},
  journal={arXiv preprint arXiv:2303.08774},
  year={2023}
}

@article{Qwen-VL,
  title={Qwen-VL: A Versatile Vision-Language Model for Understanding, Localization, Text Reading, and Beyond},
  author={Bai, Jinze and Bai, Shuai and Yang, Shusheng and Wang, Shijie and Tan, Sinan and Wang, Peng and Lin, Junyang and Zhou, Chang and Zhou, Jingren},
  journal={arXiv preprint arXiv:2308.12966},
  year={2023}
}

@inproceedings{liu2024improved,
  title={Improved baselines with visual instruction tuning},
  author={Liu, Haotian and Li, Chunyuan and Li, Yuheng and Lee, Yong Jae},
  booktitle={Proceedings of the IEEE/CVF conference on computer vision and pattern recognition},
  pages={26296--26306},
  year={2024}
}

@inproceedings{lumathvista,
  title={MathVista: Evaluating Mathematical Reasoning of Foundation Models in Visual Contexts},
  author={Lu, Pan and Bansal, Hritik and Xia, Tony and Liu, Jiacheng and Li, Chunyuan and Hajishirzi, Hannaneh and Cheng, Hao and Chang, Kai-Wei and Galley, Michel and Gao, Jianfeng},
  booktitle={The Twelfth International Conference on Learning Representations},
  year={2024}
}

@inproceedings{li2024monkey,
  title={Monkey: Image resolution and text label are important things for large multi-modal models},
  author={Li, Zhang and Yang, Biao and Liu, Qiang and Ma, Zhiyin and Zhang, Shuo and Yang, Jingxu and Sun, Yabo and Liu, Yuliang and Bai, Xiang},
  booktitle={proceedings of the IEEE/CVF conference on computer vision and pattern recognition},
  pages={26763--26773},
  year={2024}
}

@inproceedings{tang2023vistext,
  title={Vistext: A benchmark for semantically rich chart captioning},
  author={Tang, Benny and Boggust, Angie and Satyanarayan, Arvind},
  booktitle={Proceedings of the 61st Annual Meeting of the Association for Computational Linguistics (Volume 1: Long Papers)},
  pages={7268--7298},
  year={2023}
}

@incollection{mccloskey1989catastrophic,
  title={Catastrophic interference in connectionist networks: The sequential learning problem},
  author={McCloskey, Michael and Cohen, Neal J},
  booktitle={Psychology of learning and motivation},
  volume={24},
  pages={109--165},
  year={1989},
  publisher={Elsevier}
}

@inproceedings{guo-etal-2025-hide,
    title = "{H}i{D}e-{LL}a{VA}: Hierarchical Decoupling for Continual Instruction Tuning of Multimodal Large Language Model",
    author = "Guo, Haiyang  and
      Zeng, Fanhu  and
      Xiang, Ziwei  and
      Zhu, Fei  and
      Wang, Da-Han  and
      Zhang, Xu-Yao  and
      Liu, Cheng-Lin",
    editor = "Che, Wanxiang  and
      Nabende, Joyce  and
      Shutova, Ekaterina  and
      Pilehvar, Mohammad Taher",
    booktitle = "Proceedings of the 63rd Annual Meeting of the Association for Computational Linguistics (Volume 1: Long Papers)",
    month = jul,
    year = "2025",
    address = "Vienna, Austria",
    publisher = "Association for Computational Linguistics",
    url = "https://aclanthology.org/2025.acl-long.666/",
    doi = "10.18653/v1/2025.acl-long.666",
    pages = "13572--13586",
    ISBN = "979-8-89176-251-0"
}

@inproceedings{zeng2025modalprompt,
  title={Modalprompt: Towards efficient multimodal continual instruction tuning with dual-modality guided prompt},
  author={Zeng, Fanhu and Zhu, Fei and Guo, Haiyang and Zhang, Xu-Yao and Liu, Cheng-Lin},
  booktitle={Proceedings of the 2025 Conference on Empirical Methods in Natural Language Processing},
  pages={12137--12152},
  year={2025}
}

@inproceedings{wang2023orthogonal,
  title={Orthogonal subspace learning for language model continual learning},
  author={Wang, Xiao and Chen, Tianze and Ge, Qiming and Xia, Han and Bao, Rong and Zheng, Rui and Zhang, Qi and Gui, Tao and Huang, Xuan-Jing},
  booktitle={Findings of the Association for Computational Linguistics: EMNLP 2023},
  pages={10658--10671},
  year={2023}
}

@article{chen2024coin,
  title={Coin: A benchmark of continual instruction tuning for multimodel large language models},
  author={Chen, Cheng and Zhu, Junchen and Luo, Xu and Shen, Heng T and Song, Jingkuan and Gao, Lianli},
  journal={Advances in Neural Information Processing Systems},
  volume={37},
  pages={57817--57840},
  year={2024}
}

@inproceedings{huai2025cl,
  title={CL-MoE: Enhancing Multimodal Large Language Model with Dual Momentum Mixture-of-Experts for Continual Visual Question Answering},
  author={Huai, Tianyu and Zhou, Jie and Wu, Xingjiao and Chen, Qin and Bai, Qingchun and Zhou, Ze and He, Liang},
  booktitle={Proceedings of the Computer Vision and Pattern Recognition Conference},
  pages={19608--19617},
  year={2025}
}

@article{aljundi2019online,
  title={Online continual learning with maximal interfered retrieval},
  author={Aljundi, Rahaf and Belilovsky, Eugene and Tuytelaars, Tinne and Charlin, Laurent and Caccia, Massimo and Lin, Min and Page-Caccia, Lucas},
  journal={Advances in neural information processing systems},
  volume={32},
  year={2019}
}

@inproceedings{
hu2022lora,
title={Lo{RA}: Low-Rank Adaptation of Large Language Models},
author={Edward J Hu and Yelong Shen and Phillip Wallis and Zeyuan Allen-Zhu and Yuanzhi Li and Shean Wang and Lu Wang and Weizhu Chen},
booktitle={International Conference on Learning Representations},
year={2022},
url={https://openreview.net/forum?id=nZeVKeeFYf9}
}

@article{jacobs1991adaptive,
  title={Adaptive mixtures of local experts},
  author={Jacobs, Robert A and Jordan, Michael I and Nowlan, Steven J and Hinton, Geoffrey E},
  journal={Neural computation},
  volume={3},
  number={1},
  pages={79--87},
  year={1991},
  publisher={MIT Press}
}

@inproceedings{shazeer2017outrageously,
  title={Outrageously Large Neural Networks: The Sparsely-Gated Mixture-of-Experts Layer},
  author={Shazeer, Noam and Mirhoseini, Azalia and Maziarz, Krzysztof and Davis, Andy and Le, Quoc and Hinton, Geoffrey and Dean, Jeff},
  booktitle={International Conference on Learning Representations},
  year={2017}
}

@article{Jiang2023Mistral7,
  title={Mistral 7B},
  author={Albert Qiaochu Jiang and Alexandre Sablayrolles and Arthur Mensch and Chris Bamford and Devendra Singh Chaplot and Diego de Las Casas and Florian Bressand and Gianna Lengyel and Guillaume Lample and Lucile Saulnier and L{\'e}lio Renard Lavaud and Marie-Anne Lachaux and Pierre Stock and Teven Le Scao and Thibaut Lavril and Thomas Wang and Timoth{\'e}e Lacroix and William El Sayed},
  journal={ArXiv},
  year={2023},
  volume={abs/2310.06825},
  url={https://api.semanticscholar.org/CorpusID:263830494}
}

@article{touvron2023llama,
  title={Llama 2: Open foundation and fine-tuned chat models},
  author={Touvron, Hugo and Martin, Louis and Stone, Kevin and Albert, Peter and Almahairi, Amjad and Babaei, Yasmine and Bashlykov, Nikolay and Batra, Soumya and Bhargava, Prajjwal and Bhosale, Shruti and others},
  journal={arXiv preprint arXiv:2307.09288},
  year={2023}
}

@article{zhu2023minigpt,
  title={Minigpt-4: Enhancing vision-language understanding with advanced large language models},
  author={Zhu, Deyao and Chen, Jun and Shen, Xiaoqian and Li, Xiang and Elhoseiny, Mohamed},
  journal={arXiv preprint arXiv:2304.10592},
  year={2023}
}

@article{zhao2025mllm,
  title={MLLM-CL: Continual Learning for Multimodal Large Language Models},
  author={Zhao, Hongbo and Zhu, Fei and Guo, Haiyang and Wang, Meng and Wang, Rundong and Meng, Gaofeng and Zhang, Zhaoxiang},
  journal={arXiv preprint arXiv:2506.05453},
  year={2025}
}

@inproceedings{hendrycks2021many,
  title={The many faces of robustness: A critical analysis of out-of-distribution generalization},
  author={Hendrycks, Dan and Basart, Steven and Mu, Norman and Kadavath, Saurav and Wang, Frank and Dorundo, Evan and Desai, Rahul and Zhu, Tyler and Parajuli, Samyak and Guo, Mike and others},
  booktitle={Proceedings of the IEEE/CVF international conference on computer vision},
  pages={8340--8349},
  year={2021}
}

@inproceedings{li2024multimodal,
  title={Multimodal arxiv: A dataset for improving scientific comprehension of large vision-language models},
  author={Li, Lei and Wang, Yuqi and Xu, Runxin and Wang, Peiyi and Feng, Xiachong and Kong, Lingpeng and Liu, Qi},
  booktitle={Proceedings of the 62nd Annual Meeting of the Association for Computational Linguistics (Volume 1: Long Papers)},
  pages={14369--14387},
  year={2024}
}

@inproceedings{gurari2018vizwiz,
  title={Vizwiz grand challenge: Answering visual questions from blind people},
  author={Gurari, Danna and Li, Qing and Stangl, Abigale J and Guo, Anhong and Lin, Chi and Grauman, Kristen and Luo, Jiebo and Bigham, Jeffrey P},
  booktitle={Proceedings of the IEEE conference on computer vision and pattern recognition},
  pages={3608--3617},
  year={2018}
}

@inproceedings{lindstrom2022clevr,
  title={CLEVR-Math: A Dataset for Compositional Language, Visual and Mathematical Reasoning},
  author={Lindstr{\"o}m, AD and Abraham, SS},
  booktitle={Proceedings of the 16th International Workshop on Neural-Symbolic Learning and Reasoning},
  volume={3212},
  pages={155--170},
  year={2022}
}

@article{lu2021iconqa,
  title={Iconqa: A new benchmark for abstract diagram understanding and visual language reasoning},
  author={Lu, Pan and Qiu, Liang and Chen, Jiaqi and Xia, Tony and Zhao, Yizhou and Zhang, Wei and Yu, Zhou and Liang, Xiaodan and Zhu, Song-Chun},
  journal={arXiv preprint arXiv:2110.13214},
  year={2021}
}

@inproceedings{plummer2015flickr30k,
  title={Flickr30k entities: Collecting region-to-phrase correspondences for richer image-to-sentence models},
  author={Plummer, Bryan A and Wang, Liwei and Cervantes, Chris M and Caicedo, Juan C and Hockenmaier, Julia and Lazebnik, Svetlana},
  booktitle={Proceedings of the IEEE international conference on computer vision},
  pages={2641--2649},
  year={2015}
}

@article{lobry2020rsvqa,
  title={RSVQA: Visual question answering for remote sensing data},
  author={Lobry, Sylvain and Marcos, Diego and Murray, Jesse and Tuia, Devis},
  journal={IEEE Transactions on Geoscience and Remote Sensing},
  volume={58},
  number={12},
  pages={8555--8566},
  year={2020},
  publisher={IEEE}
}

@article{he2020pathvqa,
  title={Pathvqa: 30000+ questions for medical visual question answering},
  author={He, Xuehai and Zhang, Yichen and Mou, Luntian and Xing, Eric and Xie, Pengtao},
  journal={arXiv preprint arXiv:2003.10286},
  year={2020}
}

@inproceedings{sima2024drivelm,
  title={Drivelm: Driving with graph visual question answering},
  author={Sima, Chonghao and Renz, Katrin and Chitta, Kashyap and Chen, Li and Zhang, Hanxue and Xie, Chengen and Bei{\ss}wenger, Jens and Luo, Ping and Geiger, Andreas and Li, Hongyang},
  booktitle={European conference on computer vision},
  pages={256--274},
  year={2024},
  organization={Springer}
}

@article{wang2023finvis,
  title={Finvis-gpt: A multimodal large language model for financial chart analysis},
  author={Wang, Ziao and Li, Yuhang and Wu, Junda and Soon, Jaehyeon and Zhang, Xiaofeng},
  journal={arXiv preprint arXiv:2308.01430},
  year={2023}
}

@inproceedings{kembhavi2016diagram,
  title={A diagram is worth a dozen images},
  author={Kembhavi, Aniruddha and Salvato, Mike and Kolve, Eric and Seo, Minjoon and Hajishirzi, Hannaneh and Farhadi, Ali},
  booktitle={European conference on computer vision},
  pages={235--251},
  year={2016},
  organization={Springer}
}

@inproceedings{guo2025sciverse,
  title={Sciverse: Unveiling the knowledge comprehension and visual reasoning of lmms on multi-modal scientific problems},
  author={Guo, Ziyu and Zhang, Renrui and Chen, Hao and Gao, Jialin and Jiang, Dongzhi and Wang, Jiaze and Heng, Pheng-Ann},
  booktitle={Findings of the Association for Computational Linguistics: ACL 2025},
  pages={19683--19704},
  year={2025}
}

@article{chang2022mapqa,
  title={Mapqa: A dataset for question answering on choropleth maps},
  author={Chang, Shuaichen and Palzer, David and Li, Jialin and Fosler-Lussier, Eric and Xiao, Ningchuan},
  journal={arXiv preprint arXiv:2211.08545},
  year={2022}
}

@inproceedings{kembhavi2017you,
  title={Are you smarter than a sixth grader? textbook question answering for multimodal machine comprehension},
  author={Kembhavi, Aniruddha and Seo, Minjoon and Schwenk, Dustin and Choi, Jonghyun and Farhadi, Ali and Hajishirzi, Hannaneh},
  booktitle={Proceedings of the IEEE Conference on Computer Vision and Pattern recognition},
  pages={4999--5007},
  year={2017}
}

@inproceedings{wang2022learning,
  title={Learning to prompt for continual learning},
  author={Wang, Zifeng and Zhang, Zizhao and Lee, Chen-Yu and Zhang, Han and Sun, Ruoxi and Ren, Xiaoqi and Su, Guolong and Perot, Vincent and Dy, Jennifer and Pfister, Tomas},
  booktitle={Proceedings of the IEEE/CVF conference on computer vision and pattern recognition},
  pages={139--149},
  year={2022}
}

@inproceedings{radford2021learning,
  title={Learning transferable visual models from natural language supervision},
  author={Radford, Alec and Kim, Jong Wook and Hallacy, Chris and Ramesh, Aditya and Goh, Gabriel and Agarwal, Sandhini and Sastry, Girish and Askell, Amanda and Mishkin, Pamela and Clark, Jack and others},
  booktitle={International conference on machine learning},
  pages={8748--8763},
  year={2021},
  organization={PmLR}
}

@inproceedings{guo2025federated,
  title={Federated continual instruction tuning},
  author={Guo, Haiyang and Zeng, Fanhu and Zhu, Fei and Liu, Wenzhuo and Wang, Da-Han and Xu, Jian and Zhang, Xu-Yao and Liu, Cheng-Lin},
  booktitle={Proceedings of the IEEE/CVF International Conference on Computer Vision},
  pages={1325--1335},
  year={2025}
}
\end{document}